\documentclass{article}
\pdfoutput=1

\usepackage{neurips_2019}

\usepackage[utf8]{inputenc} % allow utf-8 input
\usepackage[T1]{fontenc}    % use 8-bit T1 fonts
\usepackage{hyperref}       % hyperlinks
\usepackage{url}            % simple URL typesetting
\usepackage{booktabs}       % professional-quality tables
\usepackage{amsfonts}       % blackboard math symbols
\usepackage{nicefrac}       % compact symbols for 1/2, etc.
\usepackage{microtype}      % microtypography
\usepackage[draft]{todonotes}   % notes showed

\title{Transfer Learning for Algorithm Recommendation}

\centering
\author{Gean T. Pereira\textsuperscript{1}\, Moisés R. dos Santos\textsuperscript{1}\, Edesio Alcobaça\textsuperscript{1} 
\AND 
Rafael G. Mantovani\textsuperscript{2}\, André C. P. L. F. Carvalho\textsuperscript{1} \\ 
\textsuperscript{1}Department of Computer Sciences, Institute of Mathematics and Computer Sciences\\
 University of São Paulo, São Carlos, Brazil\\
 \textsuperscript{2}Computer Engineering Department,\\
 Federal Technology University of Paraná, Apucarana, Brazil\\
 {\tt\small \{geantrinpereira,mmrsantos,edesio\}@usp.br}\\
 {\tt\small rafaelmantovani@utfpr.edu.br\, andre@icmc.usp.br}
}
\usepackage[document]{ragged2e}

\begin{document}

\maketitle
\justifying
\section{Introduction}

Meta-Learning is a subarea of Machine Learning that aims to take advantage of prior knowledge to learn faster and with fewer data~\cite{hutter2019automated}. 
There are different scenarios where meta-learning can be applied, and one of the most common is algorithm recommendation, where previous experience on applying machine learning algorithms for several datasets can be used to learn which algorithm, from a set of options, would be more suitable for a new dataset~\cite{alcobacca2018dimensionality}. 
Perhaps the most popular form of meta-learning is transfer learning, which consists of transferring knowledge acquired by a machine learning algorithm in a previous learning task to increase its performance faster in another and similar task~\cite{rice1976algorithm}. 
Transfer Learning has been widely applied in a variety of complex tasks such as image classification, machine translation and, speech recognition, achieving remarkable results~\cite{wong2018transfer, perrone2018scalable, scott2018adapted, du2017hypothesis, kumagai2016learning}.
Although transfer learning is very used in traditional or base-learning, it is still unknown if it is useful in a meta-learning setup.
For that purpose, in this paper, we investigate the effects of transferring knowledge in the meta-level instead of base-level. Thus, we train a neural network on meta-datasets related to algorithm recommendation, and then using transfer learning, we reuse the knowledge learned by the neural network in other similar datasets from the same domain, to verify how transferable is the acquired meta-knowledge. 
%ok

%Meta-learning has been widely applied to many machine learning problems. One of the problems where meta-learning has been successful it the algorithm recommendation problem.
%is one of the main meta-learning problems 
%The algorithm recommendation problem can be defined as the selection of the most suitable algorithm, for a given task, from a set of 
%candidate \cite{rice1976algorithm}
%options . 
%This problem is addressed from various perspectives in the literature, usually with classic machine learning approaches \cite{lindauer2019algorithm, alcobacca2018dimensionality, el2017performance, misir2017alors, amadini2015}. 
%Transfer learning applications are widely explored to improve machine learning methods performance \cite{wong2018transfer, perrone2018scalable, scott2018adapted, du2017hypothesis, kumagai2016learning}. 
%Transfer learning aims to improve learning task performance with knowledge gained from different but related problems \cite{pan2009survey}. Therefore, this work ...
%\textbf{FALAR DA MOTIVACAO PARA USAR META-LEARNING AQUI E SE TEM ALGO DIFERENTE DOS OUTROS USOS}

% \textbf{verificar o quão transferível é o conhecimento no nivel meta, transferindo pesos de um modelo treinado em um dataset específico a outros datasets do mesmo dominio, alterando a quantidade de conhecimento transferido... }

\section{Algorithm recommendation problem and transfer learning}

\textbf{Meta-learning for algorithm recommendation:} Meta-learning, or learning to learn, uses prior knowledge from a range of tasks, algorithms and model evaluations in order to perform better, faster and more efficient when applied to previously unseen data~\cite{finn2017model}. It is different from the traditional learning or base-learning, where the process of learning to induce a model is more focused on a specific task or dataset~\cite{brazdil2008metalearning}. In the traditional machine learning setup, algorithms are trained using features already present in the datasets, whilst in meta-learning the algorithms or meta-learners use meta-features extracted from these original datasets~\cite{hutter2019automatic}.         
A common meta-learning problem is the algorithm recommendation, where given a set of problem instances $P$ from a distribution $D$, a set $A$ of algorithms and a performance measure $m$: $P$ \textsc{x} $A$ $\rightarrow$ $\mathbb{R}$, the algorithm recommendation problem consists of finding a mapping $f$: $P$ $\rightarrow$ $A$ that optimizes the expected performance measure $m$ for instances $P$ with a distribution $D$~\cite{rice1976algorithm}. Even though algorithm recommendation is a well known problem, using meta-learning to explore prior knowledge and accelerate inference is a recent and potential trend which still requires exploration~\cite{alcobacca2018dimensionality}.
%ok

\textbf{Transfer Learning:} A formal definition of transfer Learning is given in terms of domains and learning tasks. Given a source domain $D_s$ and target domain $D_t$, learning tasks $T_s$ and $T_t$, transfer learning aims to improve the performance of $T_t$ with knowledge obtained from a different but related domain $D_s$ and learning task $T_s$, where $D_s$ $\neq$ $D_t$, or $T_s$ $\neq$ $T_t$ \cite{pan2009survey, pratt1996survey}. 
%ok

\section{Experimental results}

\textbf{Setup:} The datasets used in this paper were collected from Aslib~\cite{bischl2016aslib}, a repository associated with optimization problems such as the Traveling Salesman Problem (TSP), Quantified Boolean Formula (QBF) and Propositional Satisfiability Problem (SAT). 
We used 4 meta-datasets from Aslib related to the algorithm recommendation problem, which were preprocessed before training. 
Numerical features were normalized between 0 and 1, and targets were transformed in discrete numerical values. 
Since transfer learning requires certain standardization, the number of predictive features on the datasets were selected. 
Therefore, the Select K Best method~\cite{scikit-learn} was used to select the $K$ features with the highest score for a specific metric. 
The metric used was anova f-test, and $K$ was the number of features from the dataset with fewer features.
To handle class imbalance, we used a Stratified Hold-out validation method partitioning 80\% of the original data for train and 20\% for test.
Our meta-learner was a MultiLayer Perceptron with two hidden layers optimized with Adam, Relu activation, Cross-entropy loss and He-et-al initialization. 
The neural network was trained 30 times on each configuration and mean values for accuracy and loss were extracted, as well as standard deviations. 
For the transfer learning setup, we used three configurations: (1) ``freeze'' two hidden layers, thus setting them to be not trainable and just using their weights; (2) freeze only the first hidden layer; (3) freeze no hidden layer, thus using weights to warm-start training on other datasets.  
%ok

\textbf{Results:} 
%In cases of 2HL, pre-trained models were applied to the datasets without further training of the hidden layers, except for the adjust of the output layer.
%For 1HL, only the second hidden layers were trained, and for 0HL, all hidden layers were trained, meaning the weights of the neural networks were used as warm-start for training on other datasets. 
In the headers of Table~\ref{tab_results} are the datasets where transfer learning was applied and, above, the datasets that generated the pre-trained models. 
For example, the results above CSP-2010 (first row) refers to the final test accuracy and test loss obtained after the Normal training on CSP-2010, training using pre-trained models on CSP-MZN, CSP-Minizinc-Obj and CSP-Minizinc-Time, with the three transfer learning configurations (0HL, 1HL and 2HL).
As seen for CSP-2010, using weights from the pre-trained model on CSP-MZN as a warm-start showed slightly better accuracy in comparison with Normal training, but resulted in considerably less loss.
For CSP-MZN, the best results were obtained using transfer learning with two hidden layers from the pre-trained model of CSP-Minizinc-Obj, as well as for the pre-trained model of CSP-Minizinc-Time with one hidden layer frozen. 
Regarding CSP-Minizinc-Obj, although some executions showed competitive results, transfer learning configurations did not outperformed the Normal training. 
In the case of CSP-Minizinc-Time, the pre-trained model from CSP-MZN with two hidden layers frozen showed the best results in accuracy, quite surpassing the original/normal training. 
Our preliminary results suggests that: (1) transfer learning on the meta-level appears to be feasible and useful, achieving results comparable or superior to those obtained from training on original data; (2) transfer learning can be a potential tool for evaluating how general is the meta-knowledge leveraged by meta-learning approaches.
%ok

\begin{table*}[ht]
\tiny
\centering
\caption{Summary of the results. Columns 2HL stands for two hidden layers, which means that these layers were frozen. Columns 1HL stands for one hidden layer, meaning that the first hidden layer was not trained. Columns 0HL stands for no hidden layer, which means that no layer was frozen.}
\setlength{\tabcolsep}{4pt}
\begin{tabular}{l c | c c c | c c c | c c c}
\toprule
\multicolumn{11}{c}{\textbf{CSP-2010}} \\
\toprule
         & \multicolumn{1}{c}{\textbf{Normal}} & \multicolumn{3}{c}{\textbf{CSP-MZN}}                                           & \multicolumn{3}{c}{\textbf{CSP-Minizinc-Obj}}                                  
         & \multicolumn{3}{c}{\textbf{CSP-Minizinc-Time}} \\
\midrule
         & -                                   & 0HL                      & 1HL                      & 2HL                      & 0HL                      & 1HL                      & 2HL                      
         & 0HL                      & 1HL                      & 2HL                      \\ 
\midrule         
Acc      & 0.87 $\pm$ 0.01                     & \textbf{0.88 $\pm$ 0.01} & 0.87 $\pm$ 0.01 & 0.87 $\pm$ 0.01 & 0.87 $\pm$ 0.01 & 0.86 $\pm$ 0.01 & 0.87 $\pm$ 0.01 
         & 0.87 $\pm$ 0.01 & 0.86 $\pm$ 0.01 & 0.85 $\pm$ 0.01 \\
         
Loss     & 0.64 $\pm$ 0.15 & \textbf{0.56 $\pm$ 0.12} & 0.84 $\pm$ 0.08 & 1.01 $\pm$ 0.06 & 0.65 $\pm$ 0.15 & 0.90 $\pm$ 0.04 & 1.01 $\pm$ 0.09 
         & 0.65 $\pm$ 0.17 & 1.00 $\pm$ 0.06 & 1.07 $\pm$ 0.09 \\         

%==========================================================================================
%==========================================================================================
\toprule
\multicolumn{11}{c}{\textbf{CSP-MZN}} \\
\toprule
         & \multicolumn{1}{c}{\textbf{Normal}} & \multicolumn{3}{c}{\textbf{CSP-2010}}                                           & \multicolumn{3}{c}{\textbf{CSP-Minizinc-Obj}}                                  
         & \multicolumn{3}{c}{\textbf{CSP-Minizinc-Time}} \\
\midrule
         & -                                   & 0HL                      & 1HL                      & 2HL                      & 0HL                      & 1HL                      & 2HL                      
         & 0HL                      & 1HL                      & 2HL                      \\ 
\midrule         
Acc      & 0.71 $\pm$ 0.01                     & 0.71 $\pm$ 0.01 & 0.70 $\pm$ 0.01 & 0.71 $\pm$ 0.01 & 0.71 $\pm$ 0.01 & 0.71 $\pm$ 0.01 & \textbf{0.72 $\pm$ 0.01} 
         & 0.71 $\pm$ 0.01 & \textbf{0.72 $\pm$ 0.01} & 0.71 $\pm$ 0.01 \\
         
Loss     & 1.23 $\pm$ 0.19                     & 1.27 $\pm$ 0.22 & 1.71 $\pm$ 0.07 & 1.95 $\pm$ 0.07 & 1.25 $\pm$ 0.21 & 1.71 $\pm$ 0.08 & 1.98 $\pm$ 0.08 
         & \textbf{1.18 $\pm$ 0.18} & 1.63 $\pm$ 0.11 & 1.92 $\pm$ 0.06 \\         

%==========================================================================================
%==========================================================================================
\toprule
\multicolumn{11}{c}{\textbf{CSP-Minizinc-Obj}} \\
\toprule
         & \multicolumn{1}{c}{\textbf{Normal}} & \multicolumn{3}{c}{\textbf{CSP-2010}}                                           & \multicolumn{3}{c}{\textbf{CSP-MZN}}                                  
         & \multicolumn{3}{c}{\textbf{CSP-Minizinc-Time}} \\
\midrule
         & -                        & 0HL                      & 1HL                      & 2HL                      & 0HL                      & 1HL                      & 2HL                      
         & 0HL                      & 1HL                      & 2HL                      \\ 
\midrule         
Acc      & \textbf{0.91 $\pm$ 0.02}            & 0.87 $\pm$ 0.04 & 0.90 $\pm$ 0.00 & 0.90 $\pm$ 0.00 & 0.66 $\pm$ 0.02 & 0.70 $\pm$ 0.00 & 0.70 $\pm$ 0.00 
         & 0.90 $\pm$ 0.00 & 0.90 $\pm$ 0.00 & 0.90 $\pm$ 0.00 \\
         
Loss     & \textbf{0.77 $\pm$ 0.06}                     & 1.01 $\pm$ 0.16 & 1.37 $\pm$ 0.11 & 1.55 $\pm$ 0.06 & 3.26 $\pm$ 0.06 & 3.30 $\pm$ 0.01 & 3.27 $\pm$ 0.01 
         & 0.79 $\pm$ 0.11 & 1.01 $\pm$ 0.10 & 1.37 $\pm$ 0.15 \\         

%==========================================================================================
%==========================================================================================
\toprule
\multicolumn{11}{c}{\textbf{CSP-Minizinc-Time}} \\
\toprule
         & \multicolumn{1}{c}{\textbf{Normal}}         & \multicolumn{3}{c}{\textbf{CSP-2010}}                                           & \multicolumn{3}{c}{\textbf{CSP-MZN}}                                  
         & \multicolumn{3}{c}{\textbf{CSP-Minizinc-Obj}} \\
\midrule
         & -                        & 0HL                      & 1HL                      & 2HL                      & 0HL                      & 1HL                      & 2HL                      
         & 0HL                      & 1HL                      & 2HL                      \\ 
\midrule         
Acc      & 0.65 $\pm$ 0.00 & 0.65 $\pm$ 0.01 & 0.65 $\pm$ 0.00 & 0.65 $\pm$ 0.00 & 0.67 $\pm$ 0.03 & 0.70 $\pm$ 0.00 & \textbf{0.73 $\pm$ 0.02} 
         & 0.70 $\pm$ 0.00 & 0.70 $\pm$ 0.00 & 0.70 $\pm$ 0.00 \\
         
Loss     & 4.11 $\pm$ 0.60 & 3.78 $\pm$ 0.92 & 5.24 $\pm$ 0.18 & 5.59 $\pm$ 0.04 & \textbf{3.27 $\pm$ 0.30} & 3.81 $\pm$ 0.10 & 4.04 $\pm$ 0.03 
         & 3.87 $\pm$ 0.74 & 4.54 $\pm$ 0.07 & 4.60 $\pm$ 0.02 \\         
\bottomrule
\label{tab_results}
\end{tabular}
\end{table*}

\bibliographystyle{unsrt}

\end{document}